\documentclass[12pt]{article}
\usepackage{amsmath}
\usepackage{times}
\usepackage{graphicx}
\usepackage{color}
\usepackage{multirow}
\usepackage[authoryear]{natbib}
\usepackage{rotating}
\usepackage{bbm}
\usepackage{latexsym}

\usepackage[authoryear]{natbib}
\usepackage{rotating}
\usepackage{bbm}
\usepackage{latexsym}

\makeatletter
\def\@biblabel#1{}
\makeatother \makeatletter
\renewenvironment{thebibliography}[1]
{\section*{\refname}%
\@mkboth{\MakeUppercase\refname}{\MakeUppercase\refname}%
\list{\@biblabel{\@arabic\c@enumiv}}%
{\settowidth\labelwidth{\@biblabel{#1}}%
\leftmargin\labelwidth \advance\leftmargin\labelsep
\advance\leftmargin by 2em%
\itemindent -1.0em%
\@openbib@code
\usecounter{enumiv}%
\let\p@enumiv\@empty
\renewcommand\theenumiv{\@arabic\c@enumiv}}%
\sloppy \clubpenalty4000 \@clubpenalty \clubpenalty
\widowpenalty4000%
\sfcode`\.\@m} {\def\@noitemerr
{\@latex@warning{Empty `thebibliography' environment}}%
\endlist}
\makeatother

\newcommand{\captionfonts}{\normalsize}

\makeatletter
\long\def\@makecaption#1#2{%
  \vskip\abovecaptionskip
  \sbox\@tempboxa{{\captionfonts #1: #2}}%
  \ifdim \wd\@tempboxa >\hsize
    {\captionfonts #1: #2\par}
  \else
    \hbox to\hsize{\hfil\box\@tempboxa\hfil}%
  \fi
  \vskip\belowcaptionskip}
\makeatother

\usepackage{graphicx}
\usepackage{subfigure}
\usepackage{cite}
\newtheorem{theorem}{Theorem}
\newtheorem{corollary}{Corollary}
\newtheorem{lemma}{Lemma}

\newtheorem{proposition}{Proposition}

\newtheorem{definition}{Definition}
\newtheorem{assumption}{Assumption}

\begin{document}

\hspace{13.9cm}1

\ \vspace{20mm}\\

{\LARGE Learning rates for classification with Gaussian kernels}

\ \\
{\bf \large Shao-Bo Lin$^1$  \  \ \ Jinshan
Zeng$^{2}$
  \ \ \ Xiangyu Chang$^3$} \\
{1. Department of Statistics,
Wenzhou University, Wenzhou {\rm 325035}, China.\\
2. College of Computer Information Engineering, Jiangxi Normal
University, Nanchang, {\rm 330022}, China. \footnote{ Corresponding author: J. Zeng, email:
jsh.zeng@gmail.com} \\
3. School of Managment, Xi'an Jiaotong University, Xi'an {\rm
710049}, China. }

\begin{center} {\bf Abstract} \end{center}

This paper aims at refined error analysis for binary classification
using support vector machine (SVM) with Gaussian kernel and convex
loss. Our first result shows that for some loss functions such as
the truncated quadratic loss and quadratic loss, SVM with Gaussian
kernel can reach the almost optimal learning rate, provided the
regression function is smooth. Our second result shows that, for a
large number of loss functions, under some Tsybakov  noise
assumption, if the regression function is infinitely smooth, then
SVM with Gaussian kernel can achieve  the learning rate of order
$m^{-1}$, where $m$ is the number of samples.


\thispagestyle{empty} \markboth{}{NC instructions}
\ \vspace{-0mm}\\

\section{Introduction}

Support vector machine (SVM) is by definition the Tikhonov
regularization associated with some loss function
 over a reproducing kernel Hilbert space (RKHS). Due to its clear statistical
properties \citep{Zhang2004,Blanchard2008} and fast learning rates
\citep{Steinwart2007,Tong2016}, SVM has triggered enormous research
activities in the past twenty years. Theoretical assessments  for
the feasibility of SVM have been widely studied, to just name a few,
\citep{Chen2004,Wu2005,Wang2005,Zhou2006,Cucer2007,Wu2007,Tong2008,Steinwart2008}.

As shown in \citep{Steinwart2002}, selecting a suitable kernel
facilitates the use of SVM, both in theoretical  analysis  and
practical applications. Gaussian kernel
 {is one of}
 the most important
kernels in practice, where the width of the Gaussian kernel reflects
the frequency information for a specified learning problem
\citep{Keerthi2003}. Structures as well as explicit representations
of the inner products and norms of Gaussian RKHS have been studied
in \citep{Steinwart2006,Minh2010}. Furthermore, tight   bounds of
various covering numbers for Gaussian RKHS
 were provided in \citep{Zhou2002,Zhou2003,Steinwart2007,Kuhn2011}.
{ Based on these developed bounds, fast learning rates of SVM  with
Gaussian kernel were derived in}
\citep{Ying2007,Steinwart2007,Ye2008,Xiang2009,Xiang2011,Xiang2012,Hu2011,Eberts2011,Lin2014,Lin2015}.
As a typical  example, \citep{Steinwart2007} proved that there exist
non-trivial distributions such that the   learning rate of SVM
classification with Gaussian kernel and hinge loss can reach an
order of $m^{-1}$, where $m$ is the number of samples. Similar
results were established for SVM with   quadratic loss in
\citep{Xiang2009}.

This paper aims at refined analysis for SVM classification with
convex loss and Gaussian kernel. Our first purpose is to derive
almost optimal   learning rates for  SVM classification. Our result
shows that if the regression function (see Section
\ref{Sec.algorithm} for definition) is $r$-smooth, then SVM with
Gaussian kernel and  certain loss functions, such as the
quadratic loss and truncated quadratic loss, can reach   a learning
rate of order $m^{-r/(2r+d)+\nu}$ with arbitrarily  small positive
number $\nu$. The learning rate $m^{-r/(2r+d)}$ was proved to be
optimal in the minimax sense in \citep{Yang1999} for classification
with these  loss {functions}  when the regression function is $r$
smooth.

Since the rate $m^{-r/(2r+d)}$ is always slower than $m^{-1/2}$, our
second purpose is to deduce fast learning rates of SVM with Gaussian
kernel under  additional assumptions on the regression function. In
particular, we find that if the regression function is infinitely
differentiable and the Tsybakov noise exponent \citep{Tsybakov2004}
tends to infinity, then SVM with  Gaussian kernel and loss functions
including the hinge loss, quadratic loss,   and truncated quadratic
loss can achieve an order of $m^{-1}$. This implies that there exist
non-trivial distributions such that  learning rates of SVM with
Gaussian kernel can reach the order of $m^{-1}$, which extends the
results in \citep{Steinwart2007,Xiang2009} for the hinge loss and quadratic
loss to a general case.

The rest of paper is organized as follows. Section 2 presents some
definitions and introduces the algorithm studied in this paper.
Section \ref{Sec.Main result} provides  the main results. Section
\ref{Sec.Related work} compares our results with some related work
and gives some further discussions. Section \ref{Sec.Oracle}
establishes two oracle inequalities for SVM with convex loss.
Section \ref{Sec.Proof} gives the {proofs} of the main results.

\section{Classification  with Gaussian
Kernel and Convex Loss}\label{Sec.algorithm}

In learning theory \citep{Cucer2007,Steinwart2008}, the sample
$D=\{(x_i,y_i)\}_{i=1}^{m}$ with $x\in X=[0,1]^d$ and $y\in
Y=\{-1,1\}$ are drawn independently according to  an unknown
  distribution $\rho$ on $Z:=X\times Y$. Binary
classification algorithms produce a classifier $\mathcal
C:X\rightarrow Y$, whose generalization ability is measured by  the
misclassification error
$$
      \mathcal R(\mathcal C)=\mathbf P [\mathcal C(x)\neq y]=\int_X
      \mathbf P[y\neq\mathcal C(x)|x]d\rho_X,
$$
 where $\rho_X$ is the marginal distribution of $\rho$
and $\mathbf P[y|x]$ is the conditional probability at $x\in X$. The
Bayes rule
$$
         f_c(x)=\left\{\begin{array}{cc}
         1, & \mbox{if}\ \eta(x)\geq 1/2,\\
         -1, & \mbox{otherwise,}
         \end{array}\right.
$$
  minimizes the misclassification error, where
$\eta(x)=\mathbf P[y=1|x]$ is the Bayes decision function. Since
$f_c$ is independent of the classifier $\mathcal C$, the performance
of $\mathcal C$ can be measured by the excess misclassification
error $\mathcal R(\mathcal C)-\mathcal R(f_c)$.

Given a  loss function $\phi$, denote by $\mathcal
E^\phi(f):=\int_Z\phi(yf(x))d\rho$ the generalization error with
respect to $\phi$ and by
\begin{equation}\label{definition of frho}
    f^\phi_\rho(x):=\arg\min_{t\in\mathbf
    R}\int_Y\phi(yt)d\rho(y|x)
\end{equation}
the regression function minimizing $\mathcal E^\phi(f)$. If $\phi$
is differentiable, it is easy to check that
\begin{equation}\label{relation for regression}
       \frac{\phi'(f_\rho^\phi(x))}{\phi'(-f_\rho^\phi(x))}+1=\frac1{\eta(x)}.
\end{equation}
 We are concerned with
 the hinge loss and the following twice smooth classifying loss.

\begin{definition}\label{Def:quadratic loss}
We say that $\phi:\mathbf R\rightarrow\mathbf R_+$ is a classifying
loss (function), if it is convex, differentiable at $0$ with
$\phi'(0)<0$, and the smallest zero of $\phi$ is $1$. We say that
$\phi$ is a twice smooth classifying loss, if in addition, it is
differentialable, and its derivative is continuous and satisfies
\begin{equation}\label{Lip for loss derivative}
         |\phi'(u)-\phi'(v)|   \leq L^*|u-v|,
\end{equation}
and its modulus of convexity satisfies
\begin{equation}\label{modulus of convexity}
 \delta(\epsilon):=\inf\left\{\frac{\phi(u)+\phi(v)}2-\phi\left(\frac{u+v}2\right):u,v\in\mathbf
 R, |u-v|\geq\epsilon\right\}\geq \mu\epsilon^2.
\end{equation}
\end{definition}

The classifying loss was defined in \citep{Xiang2009} and the
modulus of convexity together with condition (\ref{modulus of
convexity}) was given in \citep{Bartlett2006}. It is easy to check
that the quadratic loss and truncated quadratic loss (or 2-norm
hinge loss) are twice smooth classifying loss. It should be mentioned
that the twice smooth classifying loss is different from the loss of
quadratic type defined in \citep{Koltchinskii2010}, since the
classifying loss requiring a zero point of $\phi$, deports the well
known logistic loss, a typical loss of quadratic type,  while the
twice differentiable property of the loss of quadratic type deports
the
 truncated quadratic loss. As concrete examples of our analysis, we
are specifically interested in loss functions presented in Table
\ref{Tab.1}.   All of them are frequently used in practical
applications \citep{Bartlett2006}. The regression functions of other
twice smooth classifying loss  can be deduced from (\ref{relation
for regression}). Since the subgradient of $\phi_h$ at $u=1$ is not
unique, the regression function for $\phi_h$ is not unique too. In
Table \ref{Tab.1}, for $\eta(x)$ not close to $1/2$, we set
$f_\rho^{\phi_h}(x)\approx\mathrm{sign}(2\eta(x)-1)$ but allow
$f_\rho^{\phi_h}(x)>1$ when $\eta(x)\approx1$,
$f_\rho^{\phi_h}(x)<-1$ when $\eta(x)\approx 0$ and
$f_\rho^{\phi_h}(x)\in(-1,1)$ when $\eta(x)\approx 1/2$.

\begin{table}[!h]
       \tabcolsep 5mm
       \caption{Loss functions and regression functions}
       \begin{center}
       \begin{tabular}{l@{}ll@{}ll@{}l}
       \hline
       {Loss function}&\qquad{Mathematical representation}
       &{Regression function $f_\rho^\phi$} \\ \hline
       Quadratic &  \qquad $\phi_{q}(u)=(1-u)^2$   &$2\eta(x)-1$ \\
       Truncated Quadratic & \qquad $\phi_{tq}=(\max\{1-u,0\})^2$    &$2\eta(x)-1$\\
       Hinge     & \qquad$\phi_h(u)=\max\{1-u,0\}$& $\mathrm{sign}(2\eta(x)-1)$\\
        \hline
       \end{tabular}
       \end{center}
       \label{Tab.1}
\end{table}

  Let
$$
                G_\sigma(x,x'):= G_\sigma(x-x'):=\exp\{-\|x-x'\|_2^2/\sigma^{2}\},\
                x,x'\in X
$$
be the Gaussian kernel, where $\sigma >0$   is  the width of
$G_\sigma$ and $\|\cdot\|_2$ denotes the Euclidean norm. Denote by
$\mathcal H_\sigma$ the RKHS associated with $G_\sigma$ endowed with
the inner product $\langle\cdot,\cdot\rangle_\sigma$ and norm
$\|\cdot\|_\sigma$.
 We consider   learning rates of the following algorithm
\begin{equation}\label{Algorithm 1}
           f_{D,\sigma,\lambda}=\arg\min_{f\in \mathcal H_\sigma}
           \left\{ \frac1{m}\sum_{i=1}^m
           \phi(y_if(x_i))+\lambda\|f\|^2_{\sigma} \right\},
\end{equation}
where $\lambda>0$ is  a regularization parameter.

\section{Main Results}\label{Sec.Main result}

Our error analysis is built upon a smoothness assumption on the
regression function, which requires the following definition.

\begin{definition}
Let $r=u+v$ for some $u\in \mathbf{N}_{0}:=\{0\}\cup \mathbf{N}$ and $0<v\leq 1$.  A
function $f:X\rightarrow \mathbf{R}$ is said to be $(r,c_0)$-smooth
if for every $\alpha =(\alpha _{1},\cdots,\alpha_{d}),\alpha_{i}\in
N_{0},$ $\sum _{j=1}^{d}\alpha_{j}=u$, the partial derivatives
$\frac{\partial ^{u}f}{\partial x{_{1}}^{\alpha_{1}}...\partial
x{_{d}}^{\alpha_{d}}}$ exist and satisfy
$$
             \left|\frac{\partial^{u}f}{\partial x{_{1}}^{\alpha_{1}}\cdots\partial
             x{_{d}}^{\alpha_{d}}}(x)
             -\frac{\partial ^{u}f}{\partial x{_{1}}^{\alpha_{1}}\cdots\partial
             x{_{d}}^{\alpha_{d}}}(x')\right |\leq c_0\|x-x'\|^{v }_2.
$$
Denote by $Lip^{(r,c_0)}$ the set of all $(r,c_0)$-smooth functions.
\end{definition}

To derive the learning rate, we need the following assumption.

\begin{assumption}\label{Assumption: Smoothness}
$f_\rho^\phi\in  Lip^{(r,c_0)}$ for some $r> 0$ and $c_0>0$.
\end{assumption}

 Assumption \ref{Assumption:
Smoothness} describes the smoothness and boundedness of the
regression function. If $\phi$ is the quadratic or truncated
quadratic loss, then the smoothness of the regression function
$f_\rho^\phi$ is equivalent to the smoothness of the Bayes decision
function $\eta$. If $\phi$ is the hinge loss, $f_\rho^{\phi_h}$ is
not unique. Assumption \ref{Assumption: Smoothness} means that there
is an $f_\rho^{\phi_h}\in Lip^{(r,c_0)}$ and implies that $\{x\in
X:\eta(x)> 1/2\}$ and $\{x\in X:\eta(x)< 1/2\}$ have a strictly
positive distance, which is a bit strict.
  Hence, for SVM with hinge loss, a
preferable assumption is a geometric noise assumption introduced in
\citep[Definition 2.3]{Steinwart2007} (see also \citep[Definition
8.15]{Steinwart2008}). Presenting learning results for SVM with
hinge loss under Assumption \ref{Assumption: Smoothness} in this
paper is for the sake of completeness.

Based on Assumption \ref{Assumption: Smoothness}, we present our
first main result.

\begin{theorem}\label{Theorem: Optimal without Tsybakov}
Let $0<\delta<1$, $\phi$ be either the hinge loss or a twice smooth
classifying loss. If Assumption \ref{Assumption: Smoothness} holds,
$\lambda=m^{-1}$ and $\sigma=m^{-1/(2r+d)}$, then for arbitrary
$0<\nu<\frac{r}{2r+d}$, with confidence at least $1-\delta$, there
holds
\begin{equation}\label{theorem1}
      \mathcal R(\mathrm{sign}(f_{D,\sigma,\lambda}))-\mathcal R(f_c)
      \leq Cm^{-\frac{r}{2r+d}+\nu}\log\frac4\delta,
\end{equation}
where $C$ is a positive constant independent of $m$ or $\delta$.
\end{theorem}

With the help of the above confidence-based error estimate, we can
derive the following learning rate in expectation.

\begin{corollary}\label{corollary: expectation without Tsybakov}
Let  $\phi$   be either the hinge loss or  a  twice smooth
classifying loss. If Assumption \ref{Assumption: Smoothness} holds,
$\lambda=m^{-1}$ and $\sigma=m^{-1/(2r+d)}$, then for arbitrary
$0<\nu<\frac{r}{2r+d}$, there holds
\begin{equation}\label{corollary2}
     \mathbf E\left[\mathcal R(\mathrm{sign}(f_{D,\sigma,\lambda}))-\mathcal
      R(f_c)\right]
      \leq 6Cm^{-\frac{r}{2r+d}+\nu},
\end{equation}
\end{corollary}
where $C$ is specified in Theorem \ref{Theorem: Optimal without Tsybakov}.

Corollary \ref{corollary: expectation without Tsybakov}
{gives an upper bound} for algorithm (\ref{Algorithm 1}) with   the
hinge loss and twice smooth classifying loss  under Assumption
\ref{Assumption: Smoothness}. However, it is difficult to judge
whether
{the bound is}
tight for all these loss functions. We obtain in the following
corollary that at least for certain specified loss functions, the
error estimate in (\ref{corollary2}) is almost optimal.

\begin{corollary}\label{corollary: Optimal without Tsybakov}
Let  $\phi$ be  either the quadratic loss or truncated quadratic
loss. If  $\lambda=m^{-1}$, $\sigma=m^{-1/(2r+d)}$, then for
arbitrary $0<\nu<\frac{r}{2r+d}$, there holds
\begin{equation}\label{corollary1}
     \overline{C}m^{-\frac{r}{2r+d}}\leq \sup_{f_\rho^\phi \in Lip^{(r,c_0)}}
     \mathbf E\left[\mathcal R(\mathrm{sign}(f_{D,\sigma,\lambda}))-\mathcal
      R(f_c)\right]
      \leq  6Cm^{-\frac{r}{2r+d}+\nu},
\end{equation}
where $\overline{C}$ is a constant  independent of  $m$, and $C$ was
specified in Theorem \ref{Theorem: Optimal without Tsybakov}.
\end{corollary}

It should be mentioned that $f_c$ depends on   $f_\rho^\phi$ and the
supremum on $f_\rho^\phi$ is equivalent to maximizing $f_{c}$ on
some set of functions.  Although the learning rate derived in
(\ref{corollary1}) is almost optimal, it is always slower than
$m^{-1/2}$. We then aim at deriving fast learning rates for
algorithm (\ref{Algorithm 1}) by imposing additional conditions
{on} the
distribution $\rho$. For this purpose, we need the following
Tsybakov noise condition \citep{Tsybakov2004}.

\begin{definition}\label{Def:Tsybakov}
Let $0\leq q\leq\infty$. We say that $\rho$ satisfies the Tsybakov
noise condition with exponent $q$ if there exists a constant
$\hat{c}_q$ such that
\begin{equation}\label{Tsybakov noise}
      \rho_X(\{x\in X:|2\eta(x)-1|\leq \hat{c}_qt\})\leq
      t^q,\qquad\forall t>0.
\end{equation}
\end{definition}
To derive the fast learning rate, we need
the following assumption.

\begin{assumption}\label{Assumption:Noise}
 $\rho$ satisfies the Tsybakov noise condition with exponent $q$.
\end{assumption}

It can be found in \citep{Tsybakov2004} that Assumption
\ref{Assumption:Noise} measures the size of the set of points that
{are corrupted with} high noise in the labeling process, and
  always holds for $q=0$ with
$\hat{c}_q=1$.
   It has been adopted
in \citep{Steinwart2007,Xiang2009,Xiang2011,Tong2016} to deduce fast
learning rates for SVM with various loss functions. Noting  that
Assumption \ref{Assumption: Smoothness} reflects the smoothness of
$\eta(\cdot)$ {while Assumption \ref{Assumption:Noise}   measures
the level of critical noise,   these two assumptions are compatible
in some sense. }
 A simple example is that when $\rho_X$ is
the uniform distribution on $X=[0,1]$, $\phi$ is the quadratic loss
and $\eta(x)=\frac{1}{2}+\frac{x}4$,   $\eta(\cdot)$ and
$f_\rho^\phi(\cdot)$   satisfy Assumption \ref{Assumption:
Smoothness} with $r=\infty$ and some $c_0>0$. Furthermore, plugging
$\eta(x)=\frac{1}{2}+\frac{x}4$ into (\ref{Tsybakov noise}),
 Assumption \ref{Assumption:Noise} holds with $\hat{c}_q=\frac12$ and
$q=1$. The following two theorems show  the improved learning rates
under Assumptions \ref{Assumption: Smoothness} and
\ref{Assumption:Noise}.

\begin{theorem}\label{Theorem:  rate with  Tsybakov}
Let $0<\delta<1$ and $\phi$ be  a twice smooth classifying loss.
Under Assumptions \ref{Assumption: Smoothness} and
\ref{Assumption:Noise}, if $\lambda=m^{-1}$ and
$\sigma=m^{-1/(2r+d)}$, then for arbitrary
$0<\nu<{\frac{2r(q+1)}{(2r+d)(q+2)}}$, with confidence at least
$1-\delta$, there holds
\begin{equation}\label{theorem2}
      \mathcal R(\mathrm{sign}(f_{D,\sigma,\lambda}))-\mathcal R(f_c)
      \leq \hat{C}m^{-\frac{2r(q+1)}{(2r+d)(q+2)}+\nu}\log\frac4\delta,
\end{equation}
where $\hat{C}$ is a constant independent of $\delta$ or $m$.
\end{theorem}
It can be found in Theorem \ref{Theorem: rate with Tsybakov} and
Corollary \ref{corollary: Optimal without Tsybakov} that the upper
bound in (\ref{theorem2}) is essentially smaller than
{the lower bound in \eqref{corollary1}. This is mainly due to the use of Assumption \ref{Assumption:Noise} in Theorem
\ref{Theorem: rate with Tsybakov}.}


\begin{theorem}\label{Theorem:  rate with  Tsybakov for hinge}
Let $0<\delta<1$ and $\phi_h$ be   the hinge loss. Under Assumptions
\ref{Assumption: Smoothness} and \ref{Assumption:Noise}, if
$\sigma=m^{-\frac{q+1}{(q+2)r+(q+1)d}}$ and $\lambda=m^{-1},$   then
for arbitrary $0<\nu<\frac{(q+1)r}{(q+2)r+(q+1)d}$, with confidence at least
$1-\delta$, there holds
\begin{equation}\label{theorem3}
      \mathcal R(\mathrm{sign}(f_{D,\sigma,\lambda}))-\mathcal R(f_c)
      \leq \hat{C}_1m^{-\frac{(q+1)r}{(q+2)r+(q+1)d}+\nu}\log\frac4\delta,
\end{equation}
where $\hat{C}_1$ is a constant independent of $\delta$ or $m$.
\end{theorem}

When  $q=0$, Theorems \ref{Theorem:  rate with  Tsybakov} and
\ref{Theorem:  rate with  Tsybakov for hinge} coincide with Theorem
\ref{Theorem: Optimal without Tsybakov}. If $r=\infty$, which
implies that the approximation error {approaches to $0$,}
then the learning rates derived in Theorems \ref{Theorem:  rate with
Tsybakov} and \ref{Theorem:  rate with Tsybakov for hinge} are
$m^{-\frac{q+1}{q+2}+\nu}$. These rates coincide with the optimal
learning rates $m^{-\frac{q+1}{q+2}}$ for certain classifiers based
on empirical risk minimization in \citep{Tsybakov2004}   up to an
arbitrarily small positive number $\nu$, and are the same as those
presented in \citep[Chapter 8]{Steinwart2008} for the hinge loss.
 Based on
Theorems \ref{Theorem: rate with Tsybakov} and  \ref{Theorem:  rate
with  Tsybakov for hinge}, we can deduce the following corollary,
showing that classification with Gaussian kernel for a large number
of loss functions can reach the rate $ m^{-1}$ for nontrivial
distributions.

\begin{corollary}\label{corollary: fast Rate with Tsybakov}
  Let $0<\delta<1$, $\phi$ be either the hinge loss or a twice
  smooth classifying
  loss. If Assumptions \ref{Assumption:
Smoothness} and \ref{Assumption:Noise} hold with $r=q=\infty$,
$\lambda=m^{-1}$ and $\sigma=1$, then for arbitrary $0<\nu<1$, {with confidence at least $1-\delta$}, there
holds
\begin{equation}\label{corollary 3}
      \mathcal R(\mathrm{sign}(f_{D,\sigma,\lambda}))-\mathcal R(f_c)
      \leq \hat{C}_2m^{-1+\nu}\log\frac4\delta,
\end{equation}
where  $\hat{C}_2$ is constant  independent of  $\delta$  or $m$.
\end{corollary}

\section{Related Work and Discussion}\label{Sec.Related work}
SVM with Gaussian kernel  and convex loss is a state-of-the-art
learning strategy for tackling  regression and classification
problems. For the regression purpose,  almost optimal learning rates
of SVM with Gaussian kernel and quadratic loss were derived in
\citep{Eberts2011}.  From regression to classification, comparison
inequalities play crucial roles in analysis.  Given a classifier
$\mathrm{sign}(f)$ and some convex loss function $\phi$, the
comparison inequality in  \citep{Chen2004} showed  that
 the excess
misclassification error $\mathcal R(\mathrm{sign}(f))-\mathcal
R(f_c)$ can be bounded by means of the  generalization error
$\mathcal E^\phi(f)-\mathcal E^\phi(f_\rho^\phi)$:
\begin{equation}\label{comparison teorem1}
          \mathcal R(\mathrm{sign}(f))-\mathcal R(f_c)\leq C_\phi\sqrt{\mathcal E^\phi(f)-\mathcal
        E^\phi(f_\rho^\phi)}.
\end{equation}
Furthermore, for $\phi_h$, \citep{Zhang2004} showed that
\begin{equation}\label{comparison teorem2}
          \mathcal R(\mathrm{sign}(f))-\mathcal R(f_c)\leq  \mathcal E^{\phi_h}(f)-\mathcal
        E^{\phi_h}(f_\rho^{\phi_h}).
\end{equation}
From (\ref{comparison teorem1}),  results in \citep{Eberts2011} can
be used to derive learning rates for
  classification with Gaussian kernel and  quadratic loss.

For other loss functions, learning rates of   classification with
Gaussian kernel were deduced in
\citep{Steinwart2007,Xiang2009,Xiang2011,Xiang2012}. In particular,
\citep{Steinwart2007}  proved that there exist non-trivial
distributions (geometric noise  assumptions for the distribution and
Tsybakov noise conditions) such that learning rates of SVM with
Gaussian kernel and hinge loss can reach an order of $m^{-1}$. Using
the refined technique in approximation theory, \citep{Xiang2009}
also constructed some distributions (smoothness assumptions for the
regression function and Tsybakov noise conditions) such that
learning rates of SVM with Gaussian kernel and  quadratic loss can
  reach an order of  $m^{-1}$.
  {Moreover,}
  \citep{Xiang2009}   deduced
learning rates for SVM with Gaussian kernel and  classifying loss,
including the $s$ norm hinge loss $\phi_{sh}(u):=(\phi_h(u))^s$ with
$s>1$ and exponential hinge loss $\phi_{eh}(u):=\max\{e^{1-u}-1,0\}$
under some smoothness assumption similar to Assumption
\ref{Assumption: Smoothness}.
{When the loss function is twice differentiable,} \citep{Xiang2011}
improved \citep{Xiang2009}'s results in terms of deriving  fast
learning rates of SVM under additional Tsybakov noise conditions.
 The main tool
  is the comparison inequality under Assumption
\ref{Assumption:Noise} \citep{Bartlett2006,Xiang2011} (see also
\citep[Theorem 8.29]{Steinwart2008}), saying that for arbitrary
measurable function $f:X\rightarrow \mathbf R$, there holds
\begin{equation}\label{comparison theorem 3}
  \mathcal R(\mathrm{sign}(f))-\mathcal R(f_c)\leq
  2^{\frac{3q+4}{q+2}}(\hat{c}_q)^{-\frac{q}{q+2}}C_{\phi,1}^{-\frac{q+1}{q+2}}
  \left\{\mathcal E^\phi(f)-\mathcal
  E^\phi(f^\phi_\rho)\right\}^{\frac{q+1}{q+2}},
\end{equation}
where $C_{\phi,1}$ is a constant depending only on $\phi$. {Since
the definition of the classifying loss in \citep{Xiang2009} deports
the logistic loss and exponential loss, \citep{Xiang2012} derived
learning rates for SVM with  some loss functions  without the
smallest zero restriction in the classifying loss.  Under this
circumstance, learning rates for SVM classification with Gaussian
kernel and logistic loss  were derived in \citep{Xiang2012}.}

 Under Assumption \ref{Assumption: Smoothness},  we  derive
almost optimal learning rates for SVM with   quadratic loss and
truncated quadratic loss. The derived learning rate in
(\ref{theorem1}) is better than the rates in   \citep[Theorem
1]{Xiang2009} with $q=0$, {while} is the same as that rate derived
in \citep{Eberts2011} for the quadratic loss. Moreover, for the
hinge loss, our result in (\ref{theorem1}) is better than that in
\citep[Theorem 4]{Xiang2009}.  Furthermore, Corollary
\ref{corollary: fast Rate with Tsybakov} shows that for some
non-trivial distributions (smoothness assumptions for the regression
function and Tsybakov noise conditions), SVM with Gaussian kernel
and hinge loss or  twice smooth classifying loss can reach the
learning rate of order $ m^{-1+\nu}$ with an arbitrarily small
positive number $\nu$. Our results extend the results in
\citep{Steinwart2007} (for hinge loss) and \citep{Xiang2009} (for
quadratic loss) to a general case.   For another widely used kernel, the polynomial kernel
$K(x,x')=(1+x\cdot x')^\tau$ with $\tau\in\mathbf N$, learning rates
for SVM with  convex loss {functions} were deduced in
\citep{Zhou2006,Tong2008}. The detailed comparisons between our
paper and  \citep{Xiang2009} (XZ2009),
 \citep{Eberts2011} (ES2013),
\citep{Tong2008} (T2008) are summarized in Table \ref{Tab.2} and
Table \ref{Tab.3}.
\begin{table}[!h]
       \tabcolsep 5mm
       \caption{Learning rates under Assumption
       \ref{Assumption: Smoothness}}
       \begin{center}
       \begin{tabular}{c@{}cc@{}cc@{}cc@{}c}
       \hline
        &\qquad {$\phi_q$} &{$\phi_{tq}$} & {$\phi_h$}   \\ \hline
       XZ2009 &  \qquad $m^{-\frac{r}{2r+4d+4}}$   & $m^{-\frac{r}{2r+4d+4}}$ & $m^{-\frac{r}{2r+4d+4}}$      \\
       ES2011  & \qquad $m^{-\frac{r}{2r+d}}$ &No & No  \\
       T2008  &\qquad $m^{-\frac{r}{2r+2d+2}}$ & $m^{-\frac{r}{2r+2d+2}}$ &$m^{-\frac{r}{2r+d+1}}$ \\
      This paper & \qquad$  m^{-\frac{r}{2r+d}} $& $  m^{-\frac{r}{2r+d}} $ &$  m^{-\frac{r}{2r+d}} $\\ \hline
       \end{tabular}
       \end{center}
       \label{Tab.2}
\end{table}
\begin{table}[!h]
       \tabcolsep 5mm
       \caption{Learning rates under Assumptions
       \ref{Assumption: Smoothness} and \ref{Assumption:Noise}}
       \begin{center}
       \begin{tabular}{c@{}cc@{}cc@{}cc@{}c}
       \hline
        & \qquad{$\phi_q$} &{$\phi_{tq}$} &{$\phi_h$}  \\ \hline
       XZ2009 &  \qquad $m^{-\frac{ r(q+1)}{(q+2)(r+2d+2)}}$   & No &No      \\
       ES2013  &\qquad $m^{-\frac{2r(q+1)}{(2r+d)(q+2)}}$ &No & No  \\
       T2008  &\qquad $m^{-\frac{r(q+1)}{2(r+d+1)(q+2)}}$& $m^{-\frac{r(q+1)}{2(r+d+1)(q+2)}}$ &$m^{-\frac{r(q+1)}{(q+2)r+(d+1)(q+1)}}$\\
      This paper & \qquad $m^{-\frac{2r(q+1)}{(2r+d)(q+2)}}$& $m^{-\frac{2r(q+1)}{(2r+d)(q+2)}}$ &
      $m^{-\frac{(q+1)r}{(q+2)r+(q+1)d}}$  \\ \hline
       \end{tabular}
       \end{center}
       \label{Tab.3}
\end{table}

Besides the smoothness assumption on the regression function,
\citep{Steinwart2007} proposed a geometric noise  assumption with
exponent $\alpha$   \citep[Definition 2.3]{Steinwart2007}  to
describe the learning rates for SVM. Based on that assumption and
Assumption \ref{Assumption:Noise} in this paper, a learning rate  of
order $m^{-\frac{2\alpha(q+1)}{2\alpha(q+2)+3q+4}}$ was derived for
SVM with Gaussian kernel and   hinge loss.
 Under the same conditions
as \citep{Steinwart2007}, \citep{Tong2016} derived a learning rate
of order $m^{-\frac{\alpha(q+1)}{\alpha(q+2)+(d+1)(q+1)}}$ for SVM
with polynomial kernels and  hinge loss. As mentioned in the
previous section, Assumption \ref{Assumption: Smoothness} for the
hinge loss implies a strictly positive distance between $\{x\in
X:\eta(x)> 1/2\}$ and $\{x\in X:\eta(x)< 1/2\}$ for arbitrary $r>0$,
which implies the geometric noise assumption with $\alpha=\infty$.
Thus a learning rate of order $m^{-\frac{q+1}{q+2}}$ can be derived
for arbitrary $r>0$. Under this circumstance, the smoothness index
$r$ fails to describe the a-priori knowledge for the classification
problems and we recommend to use the geometric noise assumption in
\citep[Definition 2.3]{Steinwart2007} or \citep[Definition
8.15]{Steinwart2008} to quantify the a-priori information. The
reason of introducing Assumption \ref{Assumption: Smoothness} to
analyze the learning rate for SVM with hinge loss is for the sake of
completeness and uniformity for analysis.

In this paper, we study the learning performance of SVM with
Gaussian kernel and convex loss. The main tools are two oracle
inequalities  developed in the next section.  Such two oracle
inequalities are different from the standard result   in
\citep[Theorem 7.23]{Steinwart2008} that  is based on a very genral
oracle inequality established in \citep[Theorem
7.20]{Steinwart2008}. To be detailed, \citep[Theorem
7.23]{Steinwart2008} requires a polynomial decaying assumption on
the (weaker) $L_2$ covering number of the RKHS but does not need the
compactness of the input space or the continuity of the kernel,
while our analysis needs Assumption \ref{Assumption: Covering} in
Section \ref{Sec.Oracle}, compactness of $X$ and continuity of $K$.
It should be mentioned that Assumption \ref{Assumption: Covering}
contains the logarithmic decaying for the covering number, which
requires some non-trivial additional work.
  We believe that by using the
established oracle inequalities and approximation results in
\citep{Zhou2006,Tong2008}, similar  error analysis for the
polynomial kernel can be derived. As far as the Gaussian kernel is
concerned, our results might be derived from the approximation error
analysis in this paper and \citep[Theorem 7.23]{Steinwart2008} with
slight changes, by using the   twice smoothness property \eqref{modulus of convexity} of the loss
functions to verify conditions of \citep[Theorem
7.23]{Steinwart2008}.
 It would be  interesting to derive
learning rates for classification with online learning and Gaussian
kernel \citep{Hu2011} and classification with Gaussian kernel and
convex loss when $X$ is a lower dimensional  manifold \citep{Ye2008}
by utilizing  the approaches in this paper.

\section{Oracle Inequalities for SVM with Convex Loss}\label{Sec.Oracle}

In this section, we present two oracle inequalities for SVM with
convex loss and Mercer  kernels.    Denote by $L_{\rho_X}^2$  the
space of    $\rho_X$ square integrable functions endowed  with
  norm $\|\cdot\|_\rho$.
 Let $\mathcal H_K$ be the RKHS
associated with a Mercer kernel $K$ endowed with norm $\|\cdot\|_K$.
Define
\begin{equation}\label{Algorithm 2}
           f_{D,\lambda}=\arg\min_{f\in \mathcal H_K}
           \left\{ \frac1{m}\sum_{i=1}^m
           \phi(y_if(x_i))+\lambda\|f\|^2_{K} \right\},
\end{equation}
where $\lambda>0$ is  a regularization parameter. Our oracle
inequalities are built upon
 the following Assumption \ref{Assumption: Covering}.
\begin{assumption}\label{Assumption: Covering}
\begin{equation}\label{coverining assumption}
         \log \mathcal N(\mathcal B_{K,1},\varepsilon) \leq \mathcal
A\psi(\varepsilon),\qquad \forall \varepsilon>0,
\end{equation}
where $\mathcal A>0$, $\psi:\mathbf R_+\rightarrow\mathbf R_+$ is a
 decreasing and continuous  function, $\mathcal B_{K,R}:=\{f\in\mathcal
H_K:\|f\|_K\leq R\}$ is the $R$ ball in $\mathcal H_K$ with some
$R>0$ and $\mathcal N(\mathcal G,\varepsilon)$ denotes the covering
number of $\mathcal G\subset C(X)$ \citep{Xiang2009}.
\end{assumption}
Assumption \ref{Assumption: Covering} depicts the capacity of RKHS.
It holds for RKHS with Gaussian kernel \citep{Steinwart2007} with
$\psi(\varepsilon)=\varepsilon^{-p}$   for  arbitrary $p\in (0,2)$,
and  for RKHS with polynomial kernel \citep{Zhou2006} with
$\psi(\varepsilon)=\log\frac{R}{\varepsilon}$ for some positive
constant $R$ independent of $\varepsilon$. Under Assumption
\ref{Assumption: Covering}, we need the  following two lemmas
derived in \citep{Shi2011,Shi2013} and \citep{Wu2005} to present the
oracle inequalities.
\begin{lemma}\label{BERNSTEIN}
 Let $\xi$ be a random variable on a probability space
$Z$ with variance $\gamma_\xi^2$ satisfying $|\xi-\mathbf E\xi|\leq
M_\xi$ for some constant $M_\xi$. Then for any $0<\delta<1$, with
confidence $1-\delta$, there holds
$$
             \frac1m\sum_{i=1}^m\xi(z_i)-\mathbf
             E\xi\leq\frac{2M_\xi\log\frac1\delta}{3m}+\sqrt{\frac{2\gamma_\xi^2\log\frac1\delta}{m}}.
$$
\end{lemma}

\begin{lemma}\label{Lemma:Concentration inequality 1.1}
Let $\mathcal G$ be a set of functions on $Z$. For every
$g\in\mathcal G$, if $|g-\mathbf Eg|\leq B$ almost everywhere and
$\mathbf E(g^2)\leq c(\mathbf E g )^\alpha$ for some $B\geq 0$,
$0\leq\alpha\leq 1$ and $c\geq0$. Then for any $\varepsilon>0$,
$$
     \mathbf P\left\{\sup_{g\in\mathcal G}\frac{\mathbf
     Eg-\frac1m\sum_{i=1}^mg(z_i)}{\sqrt{(\mathbf
     Eg)^\alpha+\varepsilon^\alpha }}>\varepsilon^{1-\frac{\alpha}2}\right\}\leq\mathcal N(\mathcal
     G,\varepsilon)\exp\left\{-\frac{m\varepsilon^{2-\alpha}}{2(c+\frac13B\varepsilon^{1-\alpha})}\right\}.
$$
\end{lemma}

\subsection{Oracle inequality for SVM with twice smooth classifying loss}
We present the first oracle
 inequality, which
describes the learning performance of  SVM with a  twice smooth
classifying loss under Assumption \ref{Assumption: Covering}.

\begin{theorem}\label{Theorem:Oracle inequality for quadratic}
Let $\phi$ be a twice smooth classifying loss. Under Assumption
\ref{Assumption: Covering}, if there exist constants  $\theta>0$ and
$C_1>0$   such that
\begin{equation}\label{Oracle condition 1}
       m^{-1}\left[\mathcal A\psi\left(\frac{\sqrt{\lambda}m^{-\theta}}
       {\|\phi'\|_{C[-1,1]}\sqrt{\phi(0)}}\right)+1
       \right] \geq C_1m^{-\theta},
\end{equation}
then for arbitrary bounded $f_0^{\phi}\in\mathcal H_K$, there holds
\begin{eqnarray}\label{th oracle 1}
      &&\mathcal E^\phi(\pi f_{D, \lambda})-\mathcal E^\phi(f_\rho^\phi)
      \leq \nonumber\\
      && C_2 \left[\mathcal D^\phi(\lambda)+ \|f_0^{\phi} - f_{\rho}^{\phi}\|_{\rho}^2 +
   \frac{
       \left[\mathcal A\psi\left(\frac{\sqrt{\lambda}m^{-\theta}}
       {\|\phi'\|_{C[-1,1]}\sqrt{\phi(0)}}\right)+1
       \right]\log\frac4\delta}{m}\right],
\end{eqnarray}
where ${\mathcal D}^{\phi}(\lambda):= {\cal E}^{\phi}(f_0^{\phi}) - {\cal E}^{\phi}(f_{\rho}^{\phi}) + \lambda \|f_0^{\phi}\|_K^2$
and  $C_2$ is a constant
independent of $m$, $\lambda$ or $\delta$ whose value is specified
in the proof  and
      $$
      \pi f(x):=\left\{\begin{array}{ll}
                                1, &\mbox{if}\
                                 f(x)>1,\\
                                 f(x), & \mbox{if}\
                                 -1\leq
                                 f(x)\leq 1, \\
                                 -1,&
                                 \mbox{if}\
                                 f(x)<-1.
                                 \end{array}
                                 \right.
$$
\end{theorem}
To prove Theorem \ref{Theorem:Oracle inequality for quadratic}, we
at first prove three propositions.

\begin{proposition}\label{Proposition:Error Decomposition}
Let $f_{D,\lambda}$ be defined by (\ref{Algorithm 2}). Then for
arbitrary $f^\phi_0\in\mathcal H_K$, there holds
\begin{equation}\label{error decomposition}
      \mathcal E^\phi(\pi f_{D, \lambda})-\mathcal E^\phi(f_\rho^\phi)
      \leq
      \mathcal D^\phi(\lambda)+\mathcal S^\phi_1(D, \lambda)
      +\mathcal S^\phi_2(D, \lambda),
\end{equation}
where
\begin{equation}\label{Approximation error}
     \mathcal D^\phi( \lambda):= \mathcal E^\phi(f_0^\phi)-\mathcal
      E^\phi(f_\rho^\phi)+\lambda\|f_0^\phi\|_K^2,
\end{equation}
\begin{equation}\label{Sample error 1}
      \mathcal S^\phi_1(D, \lambda):=
      \left[\mathcal E_D^\phi(f_0^\phi)-\mathcal E_D^\phi(f_\rho^\phi)\right]
      - \left[\mathcal E^\phi(f_0^\phi)-\mathcal
      E^\phi(f_\rho^\phi)\right],
\end{equation}
\begin{equation}\label{Sample error 2}
      \mathcal S^\phi_2(D, \lambda):=
       \left[\mathcal E^\phi(\pi f_{D, \lambda})-\mathcal
      E^\phi(f_\rho^\phi)\right]-\left[\mathcal E_D^\phi(\pi f_{D, \lambda})-\mathcal
      E_D^\phi(f_\rho^\phi)\right],
\end{equation}
and $\mathcal E_D^\phi(f):=\frac1m\sum_{i=1}^m\phi(y_if(x_i))$.
\end{proposition}

{\bf Proof.} Direct computation yields
\begin{eqnarray*}
      &&\mathcal E^\phi(\pi f_{D, \lambda})-\mathcal E^\phi(f_\rho^\phi)
      \leq
      \mathcal E^\phi(f_0^\phi)-\mathcal E^\phi(f^\phi_\rho)-\mathcal
      E^\phi(f_0^\phi)+\mathcal E^\phi_D(f_0^\phi)\\
      &+&
      \mathcal E^\phi_D(\pi
      f_{D, \lambda})-\mathcal
      E^\phi_D(f_0^\phi)-\mathcal E^\phi_D(\pi f_{D, \lambda})+
      \mathcal E^\phi(\pi f_{D, \lambda}).
\end{eqnarray*}
Since $\phi$ is a classifying loss, there holds $\mathcal
E_D^\phi(\pi f_{D, \lambda})\leq \mathcal E_D^\phi( f_{D,
\lambda})$.  Then, it follows from (\ref{Algorithm 2}) that
$$
    \mathcal E^\phi_D(\pi
      f_{D, \lambda}) \leq\mathcal E^\phi_D(
      f_{D, \lambda}) +\lambda\|f_{D, \lambda}\|_K^2
      \leq \mathcal E^\phi_D(
      f_{0}^\phi) +\lambda\|f_{0}^\phi\|_K^2.
$$
Therefore,
\begin{eqnarray*}
      &&\mathcal E^\phi(\pi f_{D, \lambda})-\mathcal E^\phi(f_\rho^\phi)
      \leq
      \mathcal E^\phi(f_0^\phi)-\mathcal E^\phi(f^\phi_\rho)+\lambda\|f\|_K^2-\mathcal
      E^\phi(f_0^\phi)+\mathcal E^\phi_D(f_0^\phi)\\
      &+&
      \mathcal E^\phi(\pi f_{D, \lambda})-\mathcal E^\phi_D(\pi
      f_{D, \lambda})
      =
      \mathcal D^\phi( \lambda)+\mathcal S^\phi_1(D, \lambda)
      +\mathcal S^\phi_2(D, \lambda).
\end{eqnarray*}
This finishes the proof of Proposition \ref{Proposition:Error
Decomposition}. \hfill $\Box$

\begin{proposition}\label{Proposition: bound s1}
  For any $0<\delta<1$, if $\phi$ is a  twice smooth classifying loss, then with confidence
$1-\frac\delta2$, there holds
$$
      \mathcal S^\phi_1(D, \lambda)\leq
             \frac{8\|\phi\|_{C[-B_0,B_0]}
             \log\frac2\delta}{3m}+
             \|\phi'\|_{C[-B_0,B_0]}\|f_0^\phi-f_\rho^\phi\|_\rho\sqrt{\frac{2\log\frac2\delta}{m}},
$$
where $B_0 := \max \{\|f_0^{\phi}\|_{\infty},1\}$.
\end{proposition}

{\bf Proof.}
 Let $\xi:=\phi(yf_0^\phi(x))-\phi(yf_\rho^\phi(x))$. Since $\phi$
 is continuous, we have
  $|\xi|\leq 2\|\phi\|_{C[-B_0,B_0]}$. Hence,
$|\xi-\mathbf E\xi|\leq 4\|\phi\|_{C[-B_0,B_0]}=:M_\xi.$ Moreover,
$|y|=1$, and  the continuous differentiability of  $\phi$
show that
$$
       |\phi(yf_0^\phi(x))-\phi(yf^\phi_\rho(x))|\leq
       \|\phi'\|_{C[-B_0,B_0]}|f_0^\phi(x)-f_\rho^\phi(x)|,
$$
which implies
\begin{equation}\label{Important}
   \gamma_\xi^2\leq \mathbf E[|\phi(yf_0^\phi(x))-\phi(yf^\phi_\rho(x))|^2]\leq
   \|\phi'\|_{C[-B_0,B_0]}^2\|f_0^\phi-f_\rho^\phi\|_\rho^2.
\end{equation}
Using  Lemma \ref{BERNSTEIN} to the random variable $\xi$, we obtain
that
$$
      \mathcal S^\phi_1(D, \lambda)\leq
             \frac{8\|\phi\|_{C[-B_0,B_0]}
             \log\frac2\delta}{3m}+ \|\phi'\|_{C[-B_0,B_0]}\|f_0^\phi-f_\rho^\phi\|_\rho\sqrt{\frac{2\log\frac2\delta}{m}}
$$
holds with confidence $1-\frac\delta2$. This finishes the proof of
Proposition \ref{Proposition: bound s1}. \hfill $\Box$

\begin{proposition}\label{Proposition:oracle bound S2 for quadratic}
Let  $0<\delta<1$. Under Assumption \ref{Assumption: Covering},  if
$\phi$ is a  twice smooth classifying loss and (\ref{Oracle condition 1})
holds for some $\theta>0$ and $C_1>0$, then
 with confidence $1-\frac\delta2$, there holds
$$
      \mathcal S^\phi_2(D, \lambda)\leq  \frac12\left[\mathcal E^\phi(\pi f_{D, \lambda})-\mathcal
    E^\phi(f_\rho^\phi)\right]+ C_3
   \frac{
       \left[\mathcal A\psi\left(\frac{\sqrt{\lambda}m^{-\theta}}
       {\|\phi'\|_{C[-1,1]}\sqrt{\phi(0)}}\right)+1
       \right]\log\frac4\delta}{m},
$$
where
      $C_3$ is a constant depending on $C_1$ and $\mu$.
\end{proposition}

{\bf Proof.} Let $R=\sqrt{\phi(0)/\lambda}$.  Set
$$
             \mathcal F_R:=\{ \phi(y \cdot \pi f)-\phi(yf_\rho^\phi):f\in
             \mathcal B_{K,R}\}.
$$
For arbitrary $g\in\mathcal F_R,$  there exists  an $f\in \mathcal
B_{K,R}$ such that $g(z)=\phi(y \cdot \pi
f(x))-\phi(yf_\rho^\phi(x)), \forall z = (x,y)$. Therefore,
$$
             \mathbf E g=\mathcal E^\phi(\pi f)-\mathcal E^\phi(f_\rho^\phi)\geq0,
             \ \ \frac1m\sum_{i=1}^mg(z_i)=\mathcal E_D^\phi(\pi f)-\mathcal E_D^\phi(f_\rho^\phi).
$$
Since $\phi$ is a classifying loss, we have $|g(z)|\leq 2 \|\phi\|_{C[-1,1]}$ and
$|g(z)-\mathbf Eg|\leq 4 \|\phi\|_{C[-1,1]}.$ Furthermore, due to (\ref{modulus of
convexity}) and the continuously differentiable property of $\phi$,
it follows from Page 150 (or Lemma 7) of \citep{Bartlett2006} that
$$
     {\bf E}(g^2)\leq (2\mu)^{-1}\|\phi'\|_{C[-1,1]}{\bf E}g.
$$
Applying Lemma \ref{Lemma:Concentration inequality 1.1} with
$\alpha=1$, $B=4 \|\phi\|_{C[-1,1]}$ and $c=(2\mu)^{-1}\|\phi'\|^2_{C[-1,1]}$
 to   $\mathcal F_R$, we obtain
\begin{eqnarray*}
       &&
       \mathbf P\left\{\sup_{g\in\mathcal F_R}
       \frac{[\mathcal E^\phi(\pi f )-\mathcal E^\phi(f_\rho^\phi)]
       -[\mathcal E_D^\phi(\pi f )-\mathcal E_D^\phi(f_\rho^\phi)]}{
       \sqrt{(\mathcal E^\phi(\pi f )-\mathcal
       E^\phi(f_\rho^\phi)) +\varepsilon }}<\sqrt{\varepsilon}\right\}\\
       &\geq&
       1-\mathcal N\left(\mathcal F_R,\varepsilon\right)\exp\left\{-\frac{m\varepsilon}{
        \mu^{-1}\|\phi'\|^2_{C[-1,1]}
       +\frac{8}3\|\phi\|_{C[-1,1]}} \right\}.
\end{eqnarray*}
Since $|y|=1$, it follows from  the convexity and continuous
differentiable property of $\phi$ that   for arbitrary
$g_1,g_2\in\mathcal F_R$, there exist $f_1,f_2\in \mathcal B_{K,R}$
and $\theta_2\in[0,1]$ such that
\begin{eqnarray*}
              |g_1(z)-g_2(z)|
              &=&
              |\phi(y\pi f_1(x))-\phi(y\pi f_2(x))|
              \leq
             \|\phi'\|_{C[-1,1]}\|\pi f_1 -\pi f_2 \|_\infty\\
             &\leq&
             \|\phi'\|_{C[-1,1]}\|f_1-f_2\|_\infty.
\end{eqnarray*}
Thus, for any $\varepsilon>0$, an $\left(\frac{\varepsilon}{
\|\phi'\|_{C[-1,1]}}\right)$-covering of $\mathcal B_{K,R}$ provides
an  $\varepsilon$-covering of $\mathcal F_R$. Therefore
$$
             \mathcal N(\mathcal F_R,\varepsilon)\leq
             \mathcal N\left(\mathcal B_{K,R},\frac{\varepsilon}{ \|\phi'\|_{C[-1,1]}}\right).
$$
Due to (\ref{Algorithm 2}), we have
$$
          \|f_{D,\lambda}\|^2_K\leq \frac{\phi(0)}{\lambda}.
$$
Then,
\begin{eqnarray}\label{S2important}
       &&
       \mathbf P\left\{\sup_{\|f\|_K\leq \sqrt{\phi(0)/\lambda}}
       \frac{[\mathcal E^\phi(\pi f )-\mathcal E^\phi(f_\rho^\phi)]
       -[\mathcal E_D^\phi(\pi f )-\mathcal E_D^\phi(f_\rho^\phi)]}{
       \sqrt{(\mathcal E^\phi(\pi f )-\mathcal
       E^\phi(f_\rho^\phi)) +\varepsilon }}\leq \sqrt{\varepsilon}\right\}\nonumber\\
       &\geq&
       1-\mathcal N\left(\mathcal B_{K,1},\frac{\varepsilon\sqrt{\lambda}}{ \|\phi'\|_{C[-1,1]}\sqrt{\phi(0)}}\right)\exp\left\{-\frac{m\varepsilon}{
       \mu^{-1}\|\phi'\|^2_{C[-1,1]}
       +\frac{8}3 \|\phi\|_{C[-1,1]}}\right\}.
\end{eqnarray}
Noting  Assumption \ref{Assumption: Covering},  we can define a
function $h$ by
$$
        h(\varepsilon):=\mathcal A\psi\left(\frac{\sqrt{\lambda}\varepsilon}
       {\|\phi'\|^2_{C[-1,1]}\sqrt{\phi(0)}}\right)- \frac{m\varepsilon}{
       \mu^{-1}\|\phi'\|_{C[-1,1]}
       +\frac{8}3 \|\phi\|_{C[-1,1]}}.
$$
Since $\psi$ is   decreasing and continuous, the function
 $h:\mathbf R_+\rightarrow\mathbf R$ is
decreasing and continuous. We can  choose $\varepsilon^*$ to be the
unique positive solution to the equation
$$
              h(\varepsilon)=\log\frac\delta2.
$$
For arbitrary $\varepsilon\geq  m^{-\theta}$ with some $\theta>0$,
there holds
\begin{equation}\label{addional 1}
               h(\varepsilon)
               \leq
               \mathcal A\psi\left(\frac{\sqrt{\lambda}m^{-\theta}}
       {\|\phi'\|_{C[-1,1]}\sqrt{\phi(0)}}\right)-\frac{m\varepsilon}{
       \mu^{-1}\|\phi'\|^2_{C[-1,1]}
       +\frac{8}3 \|\phi\|_{C[-1,1]}}=:h_1(\varepsilon).
\end{equation}
It is easy to see that $h_1(\cdot)$ is also a strictly decreasing
function. Let
$$
    \varepsilon_1
    =
   \frac{( 3\|\phi'\|_{C[-1,1]}+8 \|\phi\|_{C[-1,1]} \mu)
       \left[\mathcal A\psi\left(\frac{\sqrt{\lambda}m^{-\theta}}
       {\|\phi'\|_{C[-1,1]}\sqrt{\phi(0)}}\right)
       +\log\frac2\delta\right]}{3\mu m},
$$
then $\varepsilon_1$ is the unique solution to the equation
$$
      h_1(\varepsilon)=\log\frac\delta2.
$$
This implies
$$
    \varepsilon_1
    \leq\varepsilon_2:=
     C_3
   \frac{
       \left[\mathcal A\psi\left(\frac{\sqrt{\lambda}m^{-\theta}}
       {\|\phi'\|_{C[-1,1]}\sqrt{\phi(0)}}\right)+1
       \right]\log\frac4\delta}{m},
$$
where
$$
    C_3:= \frac{(
       3\|\phi'\|^2_{C[-1,1]}+8 \|\phi\|_{C[-1,1]} \mu)}{3\mu}
    + C_1^{-1}.
$$
Since $h_1$ is decreasing, we get $h_1(\varepsilon_2)\leq
h_1(\varepsilon_1)=\log\frac\delta2.$ According to (\ref{Oracle
condition 1}), we have $\varepsilon_2\geq m^{-\theta}$, then
(\ref{addional 1}) implies
$$
     h(\varepsilon_2)\leq
      h_1(\varepsilon_2)\leq \log\frac\delta2=h(\varepsilon^*).
$$
This together with the   decreasing and continuous property of
function $h$ yields
$$
   \varepsilon^*\leq \varepsilon_2= C_3
   \frac{
       \left[\mathcal A\psi\left(\frac{\sqrt{\lambda}m^{-\theta}}
       {\|\phi'\|_{C[-1,1]}\sqrt{\phi(0)}}\right)+1
       \right]\log\frac4\delta}{m}.
$$
The above estimate together with
 (\ref{S2important}) for $\varepsilon=\varepsilon^*$ yields that
with confidence at least  $1-\delta/2$, there holds
\begin{eqnarray*}
     &&\mathcal S_2^\phi(D, \lambda)
      =
       [\mathcal E^\phi(\pi f_{D, \lambda})-\mathcal
    E^\phi(f_\rho^\phi)] -[\mathcal E_D^\phi(\pi f_{D, \lambda})-\mathcal
    E_D^\phi(f_\rho^\phi)]\nonumber\\
    &\leq&
    \sqrt{\left[\mathcal E^\phi (\pi f_{D, \lambda})-\mathcal
    E^\phi (f_\rho^\phi)\right]+ \varepsilon^* }\sqrt{\varepsilon^*}
     \leq
    \frac12\left[\mathcal E^\phi(\pi f_{D, \lambda})-\mathcal
    E^\phi(f_\rho^\phi)\right]+ \varepsilon^*\nonumber\\
    &\leq&
    \frac12\left[\mathcal E^\phi(\pi f_{D, \lambda})-\mathcal
    E^\phi(f_\rho^\phi)\right]+ C_3
   \frac{
       \left[\mathcal A\psi\left(\frac{\sqrt{\lambda}m^{-\theta}}
       {\|\phi'\|_{C[-1,1]}\sqrt{\phi(0)}}\right)+1
       \right]\log\frac4\delta}{m},
\end{eqnarray*}
where the second inequality holds for the arithmetic-mean and geometric-mean inequality.
This  finishes the
 proof of Proposition \ref{Proposition:oracle bound S2 for quadratic}. \hfill $\Box$

 {\bf Proof of Theorem \ref{Theorem:Oracle inequality for
quadratic}.} Based on Propositions \ref{Proposition:Error
Decomposition}, \ref{Proposition: bound s1} and
\ref{Proposition:oracle bound S2 for quadratic}, we have with
confidence $1-\delta$,
\begin{eqnarray*}
      &&\mathcal E^\phi(\pi f_{D, \lambda})-\mathcal E^\phi(f_\rho^\phi)
      \leq
      \frac{8\|\phi\|_{C[-B_0,B_0]}
             \log\frac2\delta}{3m}+
             \|\phi'\|_{C[-B_0,B_0]}\|f_0^\phi-f_\rho^\phi\|_\rho\sqrt{\frac{2\log\frac2\delta}{m}}\\
             &+&
             \frac12\left[\mathcal E^\phi(\pi f_{D, \lambda})-\mathcal
    E^\phi(f_\rho^\phi)\right]+ C_3
   \frac{
       \left[\mathcal A\psi\left(\frac{\sqrt{\lambda}m^{-\theta}}
       {\|\phi'\|_{C[-1,1]}\sqrt{\phi(0)}}\right)+1
       \right]\log\frac4\delta}{m}+\mathcal D^\phi(\lambda).
\end{eqnarray*}
Therefore, with confidence $1-\delta$, there holds
\begin{eqnarray*}
      &&\mathcal E^\phi(\pi f_{D, \lambda})-\mathcal E^\phi(f_\rho^\phi)
      \leq
      C_2 \left[\mathcal D^\phi(\lambda)+ \|f_0^{\phi} - f_{\rho}^{\phi}\|_{\rho}^2 +
   \frac{
       \left[\mathcal A\psi\left(\frac{\sqrt{\lambda}m^{-\theta}}
       {\|\phi'\|_{C[-1,1]}\sqrt{\phi(0)}}\right)+1
       \right]\log\frac4\delta}{m}\right],
\end{eqnarray*}
where
$$
    C_2:=
    \frac{16}3\|\phi\|_{C[-B_0,B_0]}+2\|\phi'\|^2_{C[-B_0,B_0]}+2C_3
       +2,
$$
and to get the above inequality, we also use the following inequality
\[
\|f_0^{\phi} - f_{\rho}^{\phi}\|_{\rho} \cdot \sqrt{\frac{2\log \frac{2}{\delta}}{m}} \leq \frac{1}{2}\left(\frac{2 \log \frac{2}{\delta}}{m} +  \|f_0^{\phi} - f_{\rho}^{\phi}\|_{\rho}^2 \right).
\]
 This finishes the proof of Theorem \ref{Theorem:Oracle
inequality for quadratic}.\hfill $\Box$

\subsection{Oracle inequality for SVM with hinge loss}
The next theorem is the second oracle inequality concerning the
performance of SVM with hinge loss.

\begin{theorem}\label{Theorem:oracle hinge}
Let $0<\delta<1$. Under Assumption \ref{Assumption: Covering} and
Assumption \ref{Assumption:Noise}, if there exist constants
$\theta>0$ and $C_4>0$ such that
\begin{equation}\label{oracle condition2}
  \left[\frac{\left(\mathcal A\psi\left(
        \sqrt{\lambda}m^{-\theta}\right) +1\right)}{m}\right]^\frac{q+1}{q+2}
        \geq C_4m^{-\theta},
\end{equation}
then  for arbitrary $f_0^{\phi_h}\in\mathcal H_K$, with confidence
at least $1-\delta$ there holds
\begin{eqnarray*}
    &&\mathcal E^{\phi_h}(\pi f_{D, \lambda})-\mathcal
    E^{\phi_h}(f_\rho^{\phi_h})\\
      &\leq&
      C_5\left[\mathcal E^{\phi_h}(f_0^{\phi_h})-\mathcal
      E^\phi(f_\rho^{\phi_h})+\lambda\|f_0^{\phi_h}\|_K^2+\left[\frac{\left(\mathcal A\psi\left(
        \sqrt{\lambda}m^{-\theta}\right)
        +1\right)}{m}\right]^\frac{q+1}{q+2}\log\frac4\delta\right],
\end{eqnarray*}
where  $C_5$ is a constant independent of $m$, $\delta$
or $\lambda$ whose value is specified in the proof.
\end{theorem}

To prove Theorem \ref{Theorem:oracle hinge}, we need  two
propositions.
\begin{proposition}\label{Proposition: bound s1 for hinge}
  For any $0<\delta<1$, under Assumption \ref{Assumption:Noise},
$$
      \mathcal S^{\phi_h}_1(D, \lambda)\leq \frac{8}{3m}\log\frac2\delta+\frac{q+2}{2q+2}\left(\frac{2C_6}{m}\log\frac2\delta\right)^\frac{q+1}{q+2}
      +\frac{q}{2q+2}\left[\mathcal E^{\phi_h}(f_0^{\phi_h})-\mathcal
        E^{\phi_h}(f_\rho^{\phi_h})\right]
$$
holds with confidence $1-\frac\delta2$, where $C_6$ is a constant
independent of $m$, $\lambda$ or $\delta$.
\end{proposition}

{\bf Proof.} It can be found in \citep{Steinwart2007} that under
Assumption \ref{Assumption:Noise},
  there exists
an absolute constant $C_6\geq 1$   such that
\begin{equation}\label{IMportant for hinge}
        \mathbf
        E\left\{\left[\phi_h(yf(x))-\phi_h(yf_\rho^{\phi_h}(x))\right]^2\right\}
        \leq
        C_6\left[\mathcal E^{\phi_h}(f)-\mathcal
        E^{\phi_h}(f_\rho^{\phi_h})\right]^{\frac{q}{q+1}},\ \forall f:
        X\rightarrow[-\tilde{B},\tilde{B}].
\end{equation}
 Let $\xi:=\phi_h(yf_0^{\phi_h}(x))-\phi_h(yf_\rho^{\phi_h}(x))$ on
$(Z,\rho)$. Then, $|\xi|\leq 2$ and $|\xi-\mathbf E\xi|\leq 4.$
Moreover, it follows from (\ref{IMportant for hinge}) that
$$
        \mathbf
        E\left\{\left[\phi_h(yf_0^{\phi_h}(x))-\phi_h(yf_\rho^{\phi_h}(x))\right]^2\right\}
        \leq
        C_6\left[\mathcal E^{\phi_h}(f_0^{\phi_h})-\mathcal
        E^{\phi_h}(f_\rho^{\phi_h})\right]^{\frac{q}{q+1}}.
$$
Then, Lemma \ref{BERNSTEIN} with $\gamma_\xi^2\leq C_6\left[\mathcal
E^{\phi_h}(f_0^{\phi_h})-\mathcal
        E^{\phi_h}(f_\rho^{\phi_h})\right]^{\frac{q}{q+1}}$ and
        $M_\xi=4$ implies
that with confidence at least $1-\frac\delta2$, there holds
\begin{eqnarray*}
      \mathcal S_1(D,\sigma,\lambda)
      &\leq&
      \frac{8}{3m}\log\frac2\delta
      +\sqrt{\frac{2C_6}{m}\log\frac2\delta\left[\mathcal E^{\phi_h}(f_0^{\phi_h})-\mathcal
        E^{\phi_h}(f_\rho^{\phi_h})\right]^{\frac{q}{q+1}}}\\
        &\leq&
      \frac{8}{3m}\log\frac2\delta+\frac{q+2}{2q+2}\left(\frac{2C_6}{m}\log\frac2\delta\right)^\frac{q+1}{q+2}
      +\frac{q}{2q+2}\left[\mathcal E^{\phi_h}(f_0^{\phi_h})-\mathcal
        E^{\phi_h}(f_\rho^{\phi_h})\right],
\end{eqnarray*}
where we use the Young's inequality in the last inequality, that is,
$ab \leq \frac{a^{p_1}}{p_1} + \frac{b^{p_2}}{p_2}$ with $a = \left(\frac{2C_6}{m} \log \frac{2}{\delta}\right)^{1/2}$, $b = \left[\mathcal E^{\phi_h}(f_0^{\phi_h})-\mathcal E^{\phi_h}(f_\rho^{\phi_h})\right]^{\frac{q}{2(q+1)}}$, $p_1 = \frac{2(q+1)}{q+2}$ and $p_2 = \frac{2(q+1)}{q}$.
This
finishes the proof of Proposition \ref{Proposition: bound s1 for
hinge}. \hfill $\Box$

To  bound $\mathcal S_2^{\phi_h}(D,\sigma,\lambda)$, we need  the
 following lemma presented in \citep[Lemma 4.2]{Tong2008}.

\begin{lemma}\label{Lemma:number theory}
Let $c_1, c_2>0$ and $s>t>0$. Then the equation
$$
     x^s-c_1x^t-c_2=0
$$
has a unique positive zero $x^*$. In addition
$$
   x^*\leq\max\left\{(2c_1)^\frac1{s-t},(2c_2)^\frac1s\right\}.
$$
\end{lemma}

\begin{proposition}\label{Proposition:bound S2 for hinge oracle}
Let   $0<\delta<1$. Under Assumptions \ref{Assumption:Noise} and
\ref{Assumption: Covering}, if (\ref{oracle condition2}) holds, then
with confidence at least $1-\delta/2$, there holds
$$
      \mathcal S^{\phi_h}_2(D,\lambda)\leq C_q\left\{\mathcal E^{\phi_h}(\pi
    f_{D, \lambda})-\mathcal E^{\phi_h}(f_\rho^{\phi_h})\right\}
    +  C_7\left[\frac{\left(\mathcal A\psi\left(
        \sqrt{\lambda}m^{-\theta}\right)
        +1\right)}{m}\right]^\frac{q+1}{q+2}\log\frac4\delta,
$$
where $C_q: = \frac{q}{2q+2}\cdot 2^{\frac{q+1}{2q}}$, and $C_7$ is a constant independent of $m$, $\lambda$  or
$\delta$.
\end{proposition}

{\bf Proof.} According to (\ref{Algorithm 2}), we have $
          \lambda\|f\|^2_{K}\leq
           \phi(0)=1$.
Set
$$
             \mathcal F_R':=\{ \phi_h(y\pi f)-\phi_h(yf_\rho^{\phi_h}):f\in
             \mathcal B_{K,R}\}
$$
with $R=\lambda^{-1/2}$.
 Then for arbitrary $g\in\mathcal F_R',$  there exists an $f\in \mathcal B_{K,R}$ such
that $g(x)=\phi_h(y\cdot \pi f(x))-\phi_h(yf_\rho^{\phi_h}(x))$.
Therefore,
$$
             \mathbf E g =\mathcal E^{\phi_h}(\pi f)-\mathcal E^{\phi_h}(f_\rho^{\phi_h})
             \geq0,
             \ \ \frac1m\sum_{i=1}^mg(z_i)=\mathcal E_D^{\phi_h}(\pi f)-\mathcal E_D^{\phi_h}(f_\rho^{\phi_h}).
$$
By the definition of $\phi_h$, we have $
      |g(z)|\leq 2$ and $|g-\mathbf Eg|\leq 4.$
Furthermore, (\ref{IMportant for hinge}) yields
$$
        \mathbf
        E\left\{\left[\phi_h(y\cdot \pi f(x))-\phi_h(y  f_\rho^{\phi_h}(x))\right]^2\right\}
        \leq
        C_6\left[\mathcal E^{\phi_h}(\pi f)-\mathcal
        E^{\phi_h}(f_\rho^{\phi_h})\right]^{\frac{q}{q+1}}.
$$
Then Lemma \ref{Lemma:Concentration inequality 1.1} with
$\alpha=\frac{q}{q+1}$, $c=C_6$ and $B=4$ yields
\begin{equation}\label{inderect 1 for hinge oracle}
     \mathbf P\left\{\sup_{g\in\mathcal F'_R}\frac{\mathbf
     Eg-\frac1m\sum_{i=1}^mg(z_i)}{\sqrt{(\mathbf
     Eg)^\frac{q}{q+1}+\varepsilon^\frac{q}{q+1} }}>\varepsilon^{\frac{q+2}{2q+2}}\right\}\leq\mathcal N(\mathcal
     F'_R,\varepsilon)\exp\left\{-\frac{m\varepsilon^{\frac{q+2}{q+1}}}{2(C_6+\frac43\varepsilon^{\frac1{q+1}})}\right\}.
\end{equation}
Observe that for any $f_1,f_2\in \mathcal B_{K,R}$,
\begin{eqnarray*}
      && |(\phi_h(y\pi f_1(x))-\phi_h(y f_\rho^{\phi_h}(x)))-(\phi_h(y\cdot\pi f_2(x))-\phi_h(y
       f_\rho^{\phi_h}(x)))|\\
       &=&
       | \phi_h(y\cdot\pi f_1(x))- \phi_h(y\cdot\pi f_2(x))|
        \leq |\pi f_1(x)-\pi f_2(x)|\leq \|f_1-f_2\|_\infty.
\end{eqnarray*}
We have from $R=\lambda^{-1/2}$ that
$$
    \mathcal N(\mathcal
     F_R',\varepsilon)\leq
     \mathcal N(\mathcal B_{K,R},\varepsilon)
     \leq
     \mathcal N(\mathcal B_{K,1},\varepsilon\sqrt{\lambda}).
$$
Inserting the above estimate into (\ref{inderect 1 for hinge
oracle}), we obtain
\begin{equation}\label{inderect 2 for hinge oracle}
     \mathbf P\left\{\sup_{\|f\|_K\leq \lambda^{-1/2}}\frac{\mathbf
     Eg-\frac1m\sum_{i=1}^mg(z_i)}{\sqrt{(\mathbf
     E g)^\frac{q}{q+1}+\varepsilon^\frac{q}{q+1} }}>\varepsilon^{\frac{q+2}{2q+2}}\right\}
     \leq\mathcal N(\mathcal B_{K,1},\varepsilon\sqrt{\lambda})
     \exp\left\{-\frac{m\varepsilon^{\frac{q+2}{q+1}}}{2(C_6+\frac43\varepsilon^{\frac1{q+1}})}\right\}.
\end{equation}
According to Assumption \ref{Assumption: Covering}, we can define a
function $l:\mathbf R_+\rightarrow\mathbf R$ by
$$
        l(\varepsilon):=\mathcal A\psi\left(
        \sqrt{\lambda}\varepsilon\right)
        -\frac{m\varepsilon^{\frac{q+2}{q+1}}}{2(C_6+\frac43\varepsilon^{\frac1{q+1}})}.
$$
Since $\psi$ is decreasing, we obtain that $l(\cdot)$ is decreasing.
Thus, there exists a unique solution $\beta^*$ to the equation
$$
              l(\beta)=\log\frac\delta2.
$$
For arbitrary $\beta\geq  m^{-\theta}$ with some $\theta>0$. We have
\begin{equation}\label{additional 2 oracle}
               l(\beta)
               \leq
               \mathcal A\psi\left(
        \sqrt{\lambda}m^{-\theta}\right)
        -\frac{m\beta^{\frac{q+2}{q+1}}}{2(C_6+\frac43\beta^\frac1{q+1})}=:l_1(\beta).
\end{equation}
Take $\beta_1$ to be the positive number satisfying
$$
            l_1(\beta_1)
        =\log\frac\delta2.
$$
Then
\begin{eqnarray*}
    \beta_1^{\frac{q+2}{q+1}}-\frac{8\left(\mathcal A\psi\left(
        \sqrt{\lambda}m^{-\theta}\right)
        +\log\frac2\delta\right)}{3m}\beta_1^{\frac1{q+1}}
         - \frac{2C_6\left(\mathcal A\psi\left(
        \sqrt{\lambda}m^{-\theta}\right) +\log\frac2\delta\right)}{m}=0.
\end{eqnarray*}
Using Lemma \ref{Lemma:number theory} with $s=\frac{q+2}{q+1}$,
$t=\frac1{q+1}$,
$$
       c_1=\frac{8\left(\mathcal A\psi\left(
        \sqrt{\lambda}m^{-\theta}\right) +\log\frac2\delta\right)}{3m},
$$
and
$$
    c_2=\frac{2C_6\left(\mathcal A\psi\left(
        \sqrt{\lambda}m^{-\theta}\right) +\log\frac2\delta\right)}{m},
$$
we get
\begin{equation}\label{estimate beta 1 oracle}
   \beta_1\leq (6+4C_6)\left[\frac{\left(\mathcal A\psi\left(
        \sqrt{\lambda}m^{-\theta}\right) +\log\frac2\delta\right)}{m}\right]^\frac{q+1}{q+2}.
\end{equation}
Setting $
         C_7:=(6+4C_6)+C_4^{-1}
$, we obtain
$$
    \beta_1\leq
      C_7\left[\frac{\left(\mathcal A\psi\left(
        \sqrt{\lambda}m^{-\theta}\right) +1\right)}{m}\right]^\frac{q+1}{q+2}\log\frac4\delta
        =:\beta_2.
$$
According to (\ref{oracle condition2}), we have  $
     \beta_2\geq m^{-\theta}.
$ Then (\ref{additional 2 oracle}) implies that
$$
     l(\beta_2)\leq l_1(\beta_2)\leq
     l_1(\beta_1)=\log\frac\delta2=l(\beta^*).
$$
Hence the monotonous decreasing property of $l(\cdot)$ yields $
\beta^*\leq\beta_2$. The above estimate together with (\ref{inderect
2 for hinge oracle}) implies that with confidence at least
$1-\frac\delta2,$
\begin{eqnarray*}
     &&\mathcal S^{\phi_h}_2(D, \lambda)
      =
       [\mathcal E^{\phi_h}(\pi(f_{D, \lambda}))-\mathcal
    E^{\phi_h}(f_\rho^{\phi_h})] -[\mathcal E_D^\phi(\pi(f_{D, \lambda}))-\mathcal
    E_D^{\phi_h}(f_\rho^{\phi_h})]\\
    &\leq & \frac{q}{2q+2} \left[ (\mathcal E^{\phi_h}(\pi(f_{D, \lambda}))-\mathcal
    E^{\phi_h}(f_\rho^{\phi_h}))^{\frac{q}{q+1}} + {(\beta^*)}^{\frac{q}{q+1}} \right]^{\frac{q+1}{q}} + \frac{q+2}{2q+2} \cdot {\beta^*}\\
    &\leq&
    \frac{q}{2q+2}\cdot 2^{\frac{q+1}{2q}}\left\{\mathcal E^{\phi_h}(\pi
    f_{D, \lambda})-\mathcal E^{\phi_h}(f_\rho^{\phi_h})\right\}
    +\frac{q}{2q+2}\cdot 2^{\frac{q+1}{2q}}\beta^*+\frac{q+2}{2q+2}\beta^*\\
    &\leq&
    C_q\left\{\mathcal E^{\phi_h}(\pi
    f_{D, \lambda})-\mathcal E^{\phi_h}(f_\rho^{\phi_h})\right\}
    +   C_7\left[\frac{\left(\mathcal A\psi\left(
        \sqrt{\lambda}m^{-\theta}\right) +1\right)}{m}\right]^\frac{q+1}{q+2}\log\frac4\delta,
\end{eqnarray*}
where the first inequality holds for the Young's inequality $ab \leq \frac{a^{p_1}}{p_1} + \frac{b^{p_2}}{p_2}$ with
$a = (\mathcal E^{\phi_h}(\pi(f_{D, \lambda}))-\mathcal E^{\phi_h}(f_\rho^{\phi_h}))^{\frac{q}{q+1}} + {\varepsilon}^{\frac{q}{q+1}}$, $b = {\varepsilon}^{\frac{q+2}{2q+2}}$, $p_1 = \frac{2q+2}{q}$ and $p_2 = \frac{2q+2}{q+2}$, and the second inequality holds for the basic fact:
\[
a +b \leq \sqrt{2}(a^2+b^2)^{1/2} \leq \sqrt{2}(a^s+b^s)^{1/s}
\]
for any $1\leq s \leq 2$ and $a, b\geq 0$, and thus, $(a+b)^{\frac{q+1}{q}} \leq 2^{\frac{q+1}{2q}}(a^{\frac{q+1}{q}}+b^{\frac{q+1}{q}})$ if $q \geq 1$, where the last inequality holds for the definitions of $C_q$ and $C_7$, and the fact $2^{\frac{q+1}{2q}} \geq 1$. This
finishes the proof of   Proposition \ref{Proposition:bound S2 for
hinge oracle}.

{\bf Proof of Theorem \ref{Theorem:oracle hinge}.} Combining
Proposition \ref{Proposition:Error Decomposition} with Proposition
\ref{Proposition:bound S2 for hinge oracle} and Proposition
\ref{Proposition: bound s1 for hinge}, with confidence $1-\delta/2$,
there holds
\begin{eqnarray*}
      &&\mathcal E^{\phi_h}(\pi f_{D, \lambda})-\mathcal E^{\phi_h}(f_\rho^{\phi_h})
      \leq
      \mathcal D^{\phi_h}(\lambda)+ \frac{8}{3m}\log\frac2\delta
      +\frac{q+2}{2q+2}\left(\frac{2C_6}{m}\log\frac2\delta\right)^\frac{q+1}{q+2}\\
      &+&\frac{q}{2q+2}\left[\mathcal E^{\phi_h}(f_0^{\phi_h})-\mathcal
        E^{\phi_h}(f_\rho^{\phi_h})\right]
        +
      C_q\left\{\mathcal E^{\phi_h}(\pi
    f_{D, \lambda})-\mathcal E^{\phi_h}(f_\rho^{\phi_h})\right\}\\
    &+ & C_7\left[\frac{\left(\mathcal A\psi\left(
        \sqrt{\lambda}m^{-\theta}\right)
        +1\right)}{m}\right]^\frac{q+1}{q+2}\log\frac4\delta.
\end{eqnarray*}
Then, with confidence $1-\delta/2$, we have
$$
    \mathcal E^{\phi_h}(\pi f_{D, \lambda})-\mathcal E^{\phi_h}(f_\rho^{\phi_h})
      \leq
      C_5\left[\mathcal D^{\phi_h}(\lambda)+\left[\frac{\left(\mathcal A\psi\left(
        \sqrt{\lambda}m^{-\theta}\right)
        +1\right)}{m}\right]^\frac{q+1}{q+2}\log\frac4\delta\right],
$$
where
$$
      C_5:=(1-C_q)^{-1}\max\left\{1+\frac{q}{2q+2},\frac{8}3+\frac{C_6(q+2)}{q+1}+C_7\right\}.
$$
 The proof of Theorem \ref{Theorem:oracle hinge} is finished.
\hfill $\Box$

\section{Proofs}\label{Sec.Proof}
To prove main results in Section \ref{Sec.Main result}, we should
select an appropriate $f_0^\phi$ in
 Theorem \ref{Theorem:Oracle inequality for quadratic} and Theorem
  \ref{Theorem:oracle
hinge}. For   arbitrary $x\in [0,1]^d$, define
$F^{\phi}_{\rho,0}(x)=f^\phi_\rho(x).$ To construct a function
$F_{\rho,1}^{\phi}$   on $[-1,1]^d$, we  define
$$
      F_{\rho,1}^{\phi}(x^{(1)},\dots,x^{(j)},\dots,x^{(d)})=
      F_{\rho,0}^{\phi}(|x^{(1)}|,\dots,|x^{(j)}|,\dots,|x^{(d)}|)
$$
for $x=(x^{(1)},x^{(2)},\dots,x^{(d)})\in [-1,1]^d$ with
$j=1,2,\dots,d$. Finally, we can construct an even, continuous and
periodic function $F_\rho$ defined on $\mathbf R^d$ by
$$
            F^\phi_\rho(x^{(1)}\pm 2\ell_1,\dots,x^{(j)}\pm2\ell_j,,\dots,x^{(d)}\pm2\ell_d)=
            F_{\rho,1}^{\phi}(x^{(1)},\dots,x^{(j)},\dots,x^{(d)})
$$
with $\ell_j\in\mathbf N, j=1,\dots,d.$ We at first introduce a
kernel proposed in \citep{Eberts2011} as
$$
       K(x):=\sum_{j=1}^r\left(^r_j\right)(-1)^{1-j}\frac1{j^d}\left(\frac2{\sigma^2\pi}\right)^{d/2}
       G_{\frac{j\sigma}{\sqrt{2}}}(x),
$$
and then define
\begin{equation}\label{approximant}
            f_0^\phi(x):=K*F^\phi_\rho:=\int_{\mathbf R^d}K(x-x')F^\phi_\rho(x')dx',
           \  x\in X.
\end{equation}

 To bound the approximation error, we need the following two  lemmas
which were proved in \citep{Lin2014}.
\begin{lemma}\label{Lemma: Jackson}
     Under Assumption \ref{Assumption: Smoothness}, there holds
$$
           \|f_\rho^\phi-f_0^\phi\|_{\infty}\leq
           C_1'\sigma^r,
$$
where $C_1'$ is a constant depending only on $c_0$, $d$ and $r$.
\end{lemma}

\begin{lemma}\label{Lemma: BOUND}
 Let $f_0^\phi$ be defined by (\ref{approximant}). We have
 $f^\phi_0\in \mathcal H_\sigma$ with
$$
             \|f_0^\phi\|_\sigma\leq (\sqrt\pi)^{-d/2}(2^r-1)\sigma^{-d/2}
             \|f_\rho^\phi\|_{\infty},\
             \mbox{and}\ \
             \|f_0^\phi\|_{\infty}\leq (2^r-1)\|f_\rho^\phi\|_{\infty}.
$$
\end{lemma}

Based on the above preliminaries, we can derive the following
approximation error estimates.
\begin{proposition}\label{Proposition:approximation error}
Suppose Assumption \ref{Assumption: Smoothness} holds.
\begin{description}
\item{(a)} If $\phi$ is a  twice smooth classifying loss,  then
\begin{equation}\label{bound:diff-norm}
\|f_0^{\phi} - f_{\rho}^{\phi}\|_{\rho}^2 \leq (C_1')^2\sigma^{2r},
\end{equation}
and
\begin{equation}\label{app error for quadratic}
        \mathcal E^\phi(f_0^\phi)-\mathcal
      E^\phi(f_\rho^\phi)+\lambda\|f_0^\phi\|_\sigma^2\leq
         C_2'\left(\sigma^{2r}+\lambda\sigma^{-d}\right).
\end{equation}
\item{(b)} If $\phi_h$ is the hinge loss,  then
\begin{equation}\label{app error for hinge}
         \mathcal E^{\phi_h}(f_0^{\phi_h})-\mathcal
      E^{\phi_h}(f_\rho^{\phi_h})+\lambda\|f_0^{\phi_h}\|_\sigma^2\leq
         C_3'\left(\sigma^{r}+\lambda\sigma^{-d}\right),
\end{equation}
\end{description}
where $C_2'$  and $C_3'$ are  constants depending only on $L^*$,
$d$, $c_0$ and $r$.
\end{proposition}

{\bf Proof.} We first prove (\ref{app error for hinge}). Since
$$
   \mathcal E^{\phi_h}(f)-\mathcal
     E^{\phi_h}(f_\rho^{\phi_h})\leq
     \|f-f_\rho^{\phi_h}\|_\infty,
$$
  (\ref{app error for hinge}) follows directly from Lemmas \ref{Lemma: Jackson} and \ref{Lemma: BOUND}. Now, we turn to proving \eqref{bound:diff-norm} and (\ref{app
error for quadratic}).
\eqref{bound:diff-norm} is directly derived by Lemma \ref{Lemma: Jackson} and $\|\cdot\|_\rho\leq\|\cdot\|_\infty$.
Due to Taylor's formula, for
any $\|f\|_{\infty}\leq B_0$,
there exists an
$$
      f^*(x)\in [\min\{f(x),f^\phi_\rho(x)\},\max\{f(x),f^\phi_\rho(x)\}]
$$
such that
\begin{eqnarray*}
    &&\phi(yf(x))-\phi(yf^\phi_\rho(x))=(yf(x)-yf_\rho^{\phi}(x)) \phi'(yf^*(x))\\
    &=&
    (yf(x)-yf_\rho^{\phi}(x))\phi'(yf_\rho^\phi(x))+(yf(x)-yf_\rho^{\phi}(x))(\phi'(yf^*(x))
    -
    \phi'(yf_\rho^\phi(x)).
\end{eqnarray*}
Due to the definition of  $f_\rho^\phi$, it is easy to see
$\int_Z(yf(x)-yf_\rho^{\phi}(x))\phi'(yf_\rho^\phi(x))d\rho=0$. Hence, it
follows from (\ref{modulus of convexity}) that
$$
      \mathcal E^{\phi}(f)-\mathcal
     E^{\phi}(f_\rho^{\phi})\leq
     L^*\int_X|f(x)-f^\phi_\rho(x)||f^*(x)-f_\rho^\phi(x)|d\rho_X\leq
     L^*\|f-f_\rho^{\phi}\|_\rho^2.
$$
This together with Lemma \ref{Lemma: Jackson}, Lemma  \ref{Lemma:
BOUND} and $\|\cdot\|_\rho\leq\|\cdot\|_\infty$ yields  (\ref{app
error for quadratic}).
 This finishes the proof of
Proposition \ref{Proposition:approximation error}.\hfill $\Box$

The following covering number estimate for RKHS with Gaussian kernel
was derived in \citep[Theorem 3.1]{Steinwart2007}.

\begin{lemma}\label{LEMMA: COVERINGNUMBER}
Let $0<\sigma\leq 1$, $0<p< 2$.  There exists a constant $C_{p,d}>0$
depending only on $p$ and $d$ such that
$$
          \log \mathcal N(\mathcal B_{1},\varepsilon)\leq
          C_{p,d}\sigma^{(p/4-1)d}\varepsilon^{-p},\qquad \forall
          \varepsilon>0,
$$
where $\mathcal B_R:=\mathcal B_{R,\sigma}:=\{f\in\mathcal
H_\sigma:\|f\|_\sigma\leq R\}$.
\end{lemma}

Now, we are in a position to prove main results.

{\bf Proof of Theorem \ref{Theorem: Optimal without Tsybakov}.} For
arbitrary $0<\nu<\frac{2r}{2r+d}$, set $
       p=\frac{4(2r+d)}{12r+d}\nu.$ Then, we obtain $0<p<2$.
Plugging Proposition \ref{Proposition:approximation error} and Lemma
\ref{LEMMA: COVERINGNUMBER} into Theorem \ref{Theorem:Oracle
inequality for quadratic} with $\mathcal A=C_{p,d}\sigma^{(p/4-1)d}$
and $\psi(\varepsilon)=\varepsilon^{-p}$ and setting
$\theta=2r/(2r+d)$,  $\lambda=m^{-1}$, $\sigma=m^{-1/(2r+d)}$, and $
       p=\frac{4(2r+d)}{12r+d}\nu$, we have   (\ref{Oracle condition
1}) holds, and thus with confidence $1-\delta$, there holds
\begin{eqnarray}
\label{bound:twice-generalization}
     \mathcal E^{\phi}(\pi f_{D,\sigma, \lambda})-\mathcal
    E^{\phi}(f_\rho^{\phi})
       \leq
      \overline{C}_1\left[m^\frac{-2r}{2r+d}+
    m^{\frac{d}{2r+d}+\nu}m^{-1}
       \log\frac4\delta\right],
\end{eqnarray}
where
$$
      \overline{C}_1:=C_2\left[C_2' + (C_1')^2 +
      +C_{p,d}(\|\phi'\|_{C[-1,1]}\sqrt{\phi(0)})^{\frac{4(2r+d)}{12r+d}\nu}
      +1\right].
$$
Due to Lemma \ref{Lemma: BOUND}, we have $B_0=\max\{2^r-1,1\}$,
implying $C_2$   is bounded. Furthermore, together with the comparison inequality \eqref{comparison teorem1}, it implies
 the first part of Theorem
\ref{Theorem: Optimal without Tsybakov} for the twice smooth
classifying loss.

Then, we turn to prove Theorem \ref{Theorem: Optimal without
Tsybakov} for the hinge loss. Inserting Proposition
\ref{Proposition:approximation error} and Lemma \ref{LEMMA:
COVERINGNUMBER} into Theorem \ref{Theorem:oracle hinge} with
 $q=0$, $\mathcal A=C_{p,d}\sigma^{(p/4-1)d}$ and
$\psi(\varepsilon)=\varepsilon^{-p}$ and setting $\theta=2r/(2r+d)$,
$\lambda=m^{-1}$, $\sigma=m^{-1/(2r+d)}$, and $
       p=\frac{4(2r+d)}{12r+d}\nu,$ we have obviously (\ref{oracle
       condition2}) holds, and thus
 with confidence $1-\delta$ there holds
\begin{eqnarray*}
     \mathcal E^{{\phi_h}}(\pi f_{D,\sigma, \lambda})-\mathcal
    E^{\phi_h}(f_\rho^{\phi_h})
       \leq
      \overline{C}_2\left[m^\frac{-r}{2r+d}+
    m^{\frac{d}{2r+d}+\nu}m^{-1}
       \log\frac4\delta\right],
\end{eqnarray*}
where $
   \overline{C}_2:=C_5\left[C_3'+(C_{p,d}+1)\right].
$
This together with the comparison inequality \eqref{comparison teorem2} yields the second part of
of Theorem \ref{Theorem: Optimal
without Tsybakov} with $C:=\max\{\overline{C}_1,\overline{C}_2\}$.
\hfill $\Box$

{\bf Proof of Corollary \ref{corollary: expectation without
Tsybakov}.} From (\ref{theorem1}), we know that the nonnegative
random variable $\xi=\mathcal
R(\mathrm{sign}(f_{D,\sigma,\lambda}))-\mathcal R(f_c)$ satisfies
$$
         \mathbf P\left[\xi > t\right]
         \leq 4\exp\left\{-
         \left[Cm^{-\frac{r}{2r+d}+\nu}\right]^{-1}
         t\right\} $$
for any $t> C(\log 4) m^{-\frac{r}{2r+d}+\nu}$. Applying this bound
to the formula
$$
      \mathbf E \xi=\int_0^\infty\mathbf P\left[\xi > t\right]dt,
$$
 we obtain
\begin{eqnarray*}
        \mathbf E\left[\mathcal R(\mathrm{sign}(f_{D,\sigma,\lambda}))-\mathcal R(f_c)\right]
        &\leq& C(\log 4) m^{-\frac{r}{2r+d}+\nu} \\
        &+&  4\int_0^\infty\exp\left\{-
         \left[Cm^{-\frac{r}{2r+d}+\nu}\right]^{-1}
         t\right\}d t.
\end{eqnarray*}
By a change of variable, we see that the above integration equals
$$
      Cm^{-\frac{r}{2r+d}+\nu}\int_0^\infty   \exp\left\{-u\right\} du =Cm^{-\frac{r}{2r+d}+\nu}.
$$
Hence
$$
        \mathbf E\left[\mathcal R(\mathrm{sign}(f_{D,\sigma,\lambda}))-\mathcal R(f_c)\right]
         \leq
         6Cm^{-\frac{r}{2r+d}+\nu}.
$$
 The proof of Corollary \ref{corollary: expectation without Tsybakov} is completed. \hfill $\Box$

{\bf Proof of Corollary \ref{corollary: Optimal without Tsybakov}.}
Let $\phi$ be either the quadratic or truncated quadratic loss.
 From Table \ref{Tab.1}, we have
$\eta(x)=\frac{1+f_\rho^{\phi}(x)}2$. Hence, $f_\rho^{\phi}\in
Lip^{(r,c_0)}$ implies $\eta(\cdot)\in Lip^{(r,\frac{c_0}2)}$.  Then
the lower bound of (\ref{corollary1}) can be found in
\citep{Yang1999}, which together with (\ref{corollary2}) finishes
the proof of
 Corollary
\ref{corollary: Optimal without Tsybakov}. \hfill $\Box$

{\bf Proof of Theorem \ref{Theorem:  rate with  Tsybakov}.}
According to the comparison inequality (\ref{comparison theorem 3}),
if $\phi$ is a twice smooth classifying loss, we then obtain from
(\ref{comparison theorem 3}) and \eqref{bound:twice-generalization} that
$$
  \mathcal R(\mathrm{sign}(f))-\mathcal R(f_c)\leq
  2^{\frac{3q+4}{q+2}}(\hat{c}_q)^{-\frac{q}{q+2}}C_{\phi,1}^{-\frac{q+1}{q+2}}
  (2\bar{C}_1)^{\frac{q+1}{q+2}}m^{-\frac{2r(q+1)}{(2r+d)(q+2)}+\frac{q+1}{q+2}\nu}\log\frac4\delta.
$$
Setting
$$
     \hat{C}:=2^{\frac{3q+4}{q+2}}(\hat{c}_q)^{-\frac{q}{q+2}}C_{\phi,1}^{-\frac{q+1}{q+2}}
  (2\bar{C}_1)^{\frac{q+1}{q+2}}
$$
finishes the proof of Theorem \ref{Theorem:  rate with Tsybakov}.
\hfill $\Box$

{\bf Proof of Theorem \ref{Theorem:  rate with  Tsybakov for
hinge}.} Set $\theta=\frac{(q+1)r}{(q+1)d+(q+2)r}$,
$\lambda=m^{-1}$, $\sigma=m^{-\frac{q+1}{(q+1)d+(q+2)r}}$,  and $
       p=\frac{4[(q+2)r+(q+1)d]}{2qd+2d+6qr+8r-q-1}\nu^\frac{q+2}{q+1}.
$ Since $0<\nu<\frac{(q+1)r}{(q+2)r+(q+1)d}$, we have
$$
     0<p<\frac{4[(q+2)r+(q+1)d]}{2qd+2d+6qr+8r-q-1}\frac{(q+2)r}{(q+2)r+(q+1)d}
     <\frac{4qr+8r}{6qr+8r}<2
$$
and (\ref{oracle condition2}) holds. Then, inserting Proposition
\ref{Proposition:approximation error} and Lemma \ref{LEMMA:
COVERINGNUMBER} into Theorem \ref{Theorem:oracle hinge}, we get with
confidence $1-\delta$
\begin{eqnarray*}
      &&\mathcal E^{\phi_h}(\pi f_{D,\sigma,\lambda})-\mathcal E^{\phi_h}(f_\rho^{\phi_h})
      \leq
     \hat{C}_1m^{-\frac{(q+1)r}{(q+2)r+(q+1)d}+\nu},
\end{eqnarray*}
where $
     \hat{C}_1:=C_5(C_3'+C_{p,d}+1).
$ This finishes the proof of Theorem \ref{Theorem:  rate with
Tsybakov for hinge}. \hfill $\Box$

{\bf Proof of Corollary \ref{corollary: fast Rate with Tsybakov}}.
Since
$$
        \lim_{p\rightarrow\infty,r\rightarrow\infty}\frac{2r(q+1)}{(2r+d)(q+2)}=1
$$
and
$$
        \lim_{p\rightarrow\infty,r\rightarrow\infty}
        \frac{(q+1)r}{(q+2)r+(q+1)d}=1,
$$
(\ref{corollary 3}) follows from (\ref{theorem2}) and
(\ref{theorem3}) directly. This finishes the proof of Corollary
\ref{corollary: fast Rate with Tsybakov}. \hfill $\Box$

\section*{Acknowledgement}
We thank  Mary Ellen Perry, Ramona Marchand and Eric M Witz very much for their kindly help.
Two anonymous referees   carefully read the paper and
{gave} us numerous constructive suggestions. As a result, the
overall quality of the paper has been noticeably enhanced, to which
we feel much indebted and are grateful.  The work of S. B. Lin  and
X. Chang is supported in part by the National Natural Science
Foundation of China (Grant Nos. 61502342, 11401462). The work of J. Zeng is supported in part by the National Natural Science Foundation of China (Grants No. 61603162, 11401462) and the Doctoral start-up foundation of Jiangxi Normal University.


\end{document}